\documentclass[journal,draftcls,onecolumn]{IEEEtran}
\usepackage{cite}

\ifCLASSINFOpdf
\else
\fi
\usepackage[pdftex]{graphicx}

\usepackage{caption}
\usepackage{subcaption}

\usepackage[cmex10]{amsmath}
\usepackage{amsfonts}
\usepackage{amssymb,framed}
\usepackage{amsthm}

\usepackage{algorithmic}

\usepackage{algorithm}

\usepackage{array}
\usepackage{url}

\usepackage{color}

\usepackage{hyperref}    
\hypersetup{pdfstartview=FitH}

\newcommand{\my}[1]{{\color{black}{#1}}}
\newcommand{\myII}[1]{{\color{black}{#1}}}

\begin{document}
%
\title{Compressive hyperspectral imaging via adaptive sampling and dictionary learning}
%
%
%


\author{Mingrui~Yang,~\IEEEmembership{Member,~IEEE,}
        Frank~de~Hoog,
        ~Yuqi~Fan,
        and~Wen~Hu,~\IEEEmembership{Senior~Member,~IEEE}
        \thanks{Mingrui Yang is with the
          Department of Radiology, Case Western Reserve University,
          USA  (e-mail: mingrui.yang@case.edu).}
        \thanks{Frank de Hoog is with the Commonwealth
          Scientific and Industrial Research Organisation (CSIRO),
          Australia (email: frank.dehoog@csiro.au).}
        \thanks{Yuqi Fan is with the School of Information Technology
          and Electrical Engineering, University of Queensland,
          Australia (email: fanyuqi720@hotmail.com).}
        \thanks{Wen Hu is with the School of
Computer Science and Engineering, University of New South Wales, and CSIRO, Australia (email: wen.hu@unsw.edu.au).}
}

%
%

\markboth{Journal of \LaTeX\ Class Files,~Vol.~xx, No.~x, Month~Year}%
{Shell \MakeLowercase{\textit{et al.}}: Bare Demo of IEEEtran.cls for Journals}
%



\maketitle

\begin{abstract}
In this paper, we propose a new sampling strategy for hyperspectral signals
that is based on dictionary learning and singular
value decomposition (SVD). Specifically, we first learn a sparsifying
dictionary from training spectral data using dictionary learning. We
then perform an SVD on the dictionary and use the first
few left singular vectors as the rows of the measurement matrix to obtain the
compressive measurements for reconstruction. The proposed method provides significant
improvement over the conventional compressive sensing approaches. The
reconstruction performance is further improved by reconditioning the
sensing matrix using matrix balancing. We also demonstrate that
the combination of dictionary learning and SVD is robust by applying them to different datasets. 
\end{abstract}


%
\IEEEpeerreviewmaketitle

\section{Introduction}

Hyperspectral imaging (HSI) is an optical acquisition
technique that has broad applications in agriculture, mineralogy,
manufacturing, and surveillance. It captures reflectance of light and
decomposes it into hundreds or even thousands of spectral bands ranging from
ultraviolet to infrared wavelength. Compared to
conventional RGB imaging systems, HSI provides substantially more
detailed signatures and as a result, it makes identification of
more features in the subject of interest feasible, and provides
much better accuracy.

However, significantly more signatures lead to significantly larger
datasets. Hundreds or thousands of spectral bands introduce a
third dimension to the two dimensional (2D) spatial image, resulting in a
three dimensional (3D) data cube. Capturing, processing and analyzing this
hyperspectral data cube poses significant challenges. First, the
design of hyperspectral imagers has to take into account factors such
as photon efficiency, acquisition time, dynamic range, sensor
characteristics and cost. The most common technique in practice is to use a line spectrometer to scan through the
spatial dimensions, and this is the approach taken, for
instance, in remote sensing \cite{202371,685819}. Second, the
hyperspectral data cube acquired can easily reach 1GB in size,
making data transmission and analysis difficult. This
high dimensionality also limits the ability of current technology to
preprocess the data (e.g. denoising) or identify the significant
spectral signatures.

To overcome this ``curse of dimensionality'', researchers have
recently made substantial progress in applying techniques developed in
compressive sensing (CS) \cite{1580791,Do} to HSI
in order to reduce the amount of data that needs to be acquired and processed.
The basic
idea of CS is to utilize the special structures of the signal itself
(or in a transformed domain) so that a small number of measured projections can be used to reconstruct the signal. The special structures in particular refer to the sparsity or
compressibility of the signal. We say that a signal is sparse if it
(or its representation in a transform domain) has only a limited
number of nonzero elements. More precisely, it is $k$-sparse if it has
no more than $k$ nonzero elements. A signal is said to be
compressible if the signal (or its representation in a transform domain)
has an exponential decay after sorting in magnitude. Under suitable
conditions, a reconstruction can then be accomplished via $\ell_1$ minimization or even
non-convex optimization techniques.

There have been many studies in applications of CS, one of the first
applications being the well-known single pixel camera \cite{4472247},
which measures a small number of random projections of an image from
which it successfully reconstructs the image. Recently, substantial effort has been put into applying CS to hyperspectral
imaging because the additional spectral information it provides
enables a much greater range of applications, especially
for classification and pattern recognition problems. For
instance, August~et~al.\cite{August:13} have proposed adding another
stage to decompose the single-pixel light field from the single pixel
camera into different wavelength bands and use a digital micro-mirror
device (DMD) to randomly collect them. This process basically mimics the
single pixel camera, but in the spectral domain. It uses the 0-1
random binary pattern to approximate the $\pm 1$ symmetric Bernoulli
pattern (also known as the Rademacher distribution) that is often used
for subsampling in CS. The sparsifying basis for reconstruction they
use is the 3D Haar wavelet transform. More recently, Zhang et al.~\cite{6522858} and
Zhao and Yang~\cite{6819824} used more flexible approaches based on
specific sparse representation dictionaries to perform HSI classification and
denoising. 

\my{
  In recent years, a number of researchers have proposed learning appropriate dictionaries rather than using an a priori choice. For example, Charles~et~al.\cite{5762314} applied dictionary learning to
HSI data and showed that the learned dictionary improved the performance
of a supervised classification algorithm compared to a dictionary with
predefined endmembers. In another recent paper,
Lin~et~al. \cite{Lin:2014:SEC:2661229.2661262} proposed the use of a new
sparse representation of spectral signals based on dictionary
learning, and combined it with the 0-1 random binary sensing matrix
to encode and decode the HS images.
}Hahn et al.~\cite{6853751} adopted the idea of dictionary
  learning to learn a sparsifying near orthogonal basis that did not
  explicitly represent the endmembers. They then applied
  eigenvalue decomposition to find a near optimal adaptive measurement
  matrix and improved classification accuracy using the compressed data.

In this paper, we build on the work in~\cite{5762314,6853751,6847688} and propose a
new sampling strategy based on a sparsifying dictionary of the spectral
signals learned from dictionary learning. Specifically, we first learn
a sparsifying dictionary from the spectral training data using sparse
coding. We then perform a singular value decomposition (SVD) on the
learned dictionary and use the first $m$ left singular vectors (corresponding to singular values in
descending order) as the rows of the measurement matrix to obtain the $m$
compressive measurements. These compressive measurements can then be used
for reconstruction using standard CS decoding techniques. The
reconstruction performance is further improved by reconditioning the
sensing matrix, where a matrix balancing technique is
employed. \my{Moreover, we show by experiment that the dictionary learned
from a subset of the training data and the corresponding singular
vectors are robust when applied to the test data from a different scene.}

\my{
  In particular, we have made the following technical contributions:

\begin{itemize}
  \item We have made substantial improvements over the established CS based
    methods for HSI. We first learn a
    sparsifying dictionary by dictionary learning and then use a small
    number of its left singular vectors as the measurement
    matrix. Reconstruction is then performed using standard CS
    methods. Experiments demonstrate that this approach performs significantly
    better than the conventional CS methods.
  \item We have proposed a new matrix balancing algorithm. It improves
    the condition of the sensing matrix substantially, leading to a
    further improvement in the performance of our proposed sampling strategy.
  \item We have also investigated the robustness of our proposed method and
    shown, by experiment, that a dictionary learned from a given scene can be applied to different, but similar scenes.
  \end{itemize}
  
  The rest of the paper is organized as follows. In
  Section~\ref{sec:background}, we first introduce briefly the
  background of the established methods, including CS and
  dictionary learning. The technical details for adaptive sampling
  using SVD, and the matrix balancing algorithm are presented in
  Section~\ref{sec:methods}. 
  Our proposed methods, including the
  robustness of the adaptive sampling, are then evaluated in
  Section~\ref{sec:evaluation}. Finally, Section~\ref{sec:conclusion}
  gives some concluding remarks.
}

\section{Background}\label{sec:background}

In the following sections, we give a brief introduction to conventional compressive sensing and dictionary learning.

\subsection{Background for Compressive Sensing}

About one decade ago, Candes, Romberg, and Tao~\cite{1580791}, and
independently, Donoho~\cite{Do} made an important breakthrough in the
sampling theory, which they named Compressive
Sensing, or Compressed Sensing (CS). The breakthrough is that CS
gives, under certain conditions, a theoretical guarantee
that one can sample a continuous signal at a rate much smaller than
the Nyquist rate but
still is able to recover the original signal
accurately. Specifically, the CS problem can be formulated as follows. First, a signal $x\in \mathbb{R}^n$ is $k$-sparse if
$\|x\|_0 \le k$, where the $\ell_0$ ``norm'' simply counts the number
of nonzeros in $x$. Now let $x \in \mathbb{R}^n$ be the discretized original
signal of interest, which is $k$-sparse in a certain basis $\Psi \in
\mathbb{R}^{n \times n}$ (i.e.,
$\|s\|_0 \le k$ where $x = \Psi s$). Then the compressive measurements
can be obtained by different random linear combinations
of the elements of $x$, which results in the model $y = \Phi x$,
where $\Phi \in \mathbb{R}^{m\times n}$ and $y \in \mathbb{R}^m$ with
$m \ll n$. It is known that if the elements of the measurement matrix
$\Phi$ are drawn from i.i.d. Gaussian, Rademacher, or other sub-Gaussian 
distributions, and $m$ is of order $k\log (n/k)$, then the sparse
coefficient vector $s$ (and so
$x$) can be recovered exactly with high probability by solving the
following optimization problem
\begin{align}
  \min_s \|s\|_0 \quad \text{ subject to } \quad y = As,
  \label{eqn:l0}
\end{align}
where $A = \Phi \Psi$ is the sensing matrix.\footnote{There has been ambiguity in the
 CS literature. In this paper, we call $\Phi$ the measurement matrix, and
name $A$ the sensing matrix.} However, solving this problem directly is NP
hard. Fortunately, Candes, Romberg, and Tao~\cite{1580791} have shown
that given the conditions above, solving the optimization
problem~\eqref{eqn:l0} is equivalent to solving
\begin{align}
  \min_s \|s\|_1 \quad \text{ subject to } \quad y = As,
  \label{eqn:l1}
\end{align}
which can be solved by linear programming methods in polynomial time. Moreover, if the
measurements $y$ are contaminated with noise $w$ with $\|w\|_2 < \epsilon$,
then \eqref{eqn:l1} can be formulated as a basis pursuit denoising
(BPDN) problem
\begin{align}
  \min_{\tilde{s}} \|\tilde{s}\|_1 \quad \text{subject to} \quad \|y -
  A\tilde{s}\|_2 < \epsilon.
  \label{eqn:l1_noisy}
\end{align}
It has been shown by Candes~\cite{Candes:08} that given certain conditions on matrix $A$, the solution $s^*$ to Eq.~\eqref{eqn:l1_noisy} obeys
\begin{align}
  \|s^* - s\|_2 \le C_0 \|s - s_k\|_1/\sqrt{k} + C_1\epsilon,
\end{align}
with some constants $C_0$ and $C_1$, where $s_k$ is the vector
obtained by setting all but the $k$-largest
entries of $s$  in magnitude to zero. This result shows that the CS process is
stable. In particular, the reconstruction error from CS can be bounded
by the best $k$-term approximation with variation up to the noise
level. There are many solvers in
the literature for solving the minimization
problem~\eqref{eqn:l1_noisy}. In this paper, we use the well known
SPGL1 solver~\cite{spgl1:2007,BergFriedlander:2008} to solve the BPDN
problem~\eqref{eqn:l1_noisy}.

For conventional CS theory, one of the key factors for successfully
recovering a sparse signal is that the sparsifying domain $\Psi$ is an
orthonormal basis, e.g. discrete cosine transform (DCT) or wavelets
bases. On the other hand, researchers have found overcomplete
representations, such as the Gabor
frames~\cite{Feichtinger:1997:GAA:550022}, to be extremely flexible and effective~\cite{Starck2004287,4060954}, and a view is developing
that overcomplete representations can be equally helpful in CS
problems~\cite{Candes201159}. In many other applications, a good
sparsifying orthonormal basis for a particular signal of interest is
often difficult to find~\cite{CPA:CPA10116}. In this case, dictionary
learning becomes a useful tool for finding sparsifying domains.

\subsection{Background on Dictionary Learning}

Dictionary learning is a machine learning technique to learn a
sparsifying dictionary and the corresponding sparse coefficients for a
given set of training data. It is closely connected with sparse
approximation and compressive sensing. Let $\Phi\in \mathbb{R}^{m\times d}$ denote the
measurement matrix and $D\in \mathbb{R}^{d \times n}$ denote the
sparsifying dictionary. Let $y\in \mathbb{R}^m$ be
the measurement vector obtained from applying the measurement matrix
$\Phi$ to the signal of interest $x = Ds$, where $s\in \mathbb{R}^n$ is sparse. Then the problem of
reconstructing the original signal $x\in \mathbb{R}^d$ can be
formulated as the following BPDN problem
\begin{align}
  \min_s \|s\|_1 \quad \text{s.t.} \quad \|y-As\|_2 < \epsilon,
  \label{eqn:BPDN_dictionary}
\end{align}
where $A = \Phi D$ is the sensing matrix and $\epsilon$ is the error
bound. Then $x$ can be recovered from $x = Ds$. In practice, the BPDN
problem~\eqref{eqn:BPDN_dictionary} can
also be solved
by reformulating it into a Lasso problem
\begin{align}
  \min_s \frac{1}{2}\|y - As\|_2^2 + \lambda\|s\|_1 \label{eqn:lasso}
\end{align}
where the regularization parameter $\lambda$ controls how compressible
the solution $\hat s$ will be. \myII{However, there is no analytical
  link between the value of the parameter $\lambda$ and the effective
  compressibility of $s$.}

If the sparsifying domain is unknown, the sparsifying dictionary
$D$ can be calculated from a set of training
signals $\{x_i\}_{i=1}^p$ as follows:
\begin{align}
  \min_{s_i, D} \sum_{i=1}^p \frac{1}{2} \|x_i - Ds_i\|_2^2 + \lambda
  \|s_i\|_1 \notag \\
  \text{s.t.} \quad \|d_j\|_2 \le 1 \quad \forall \; j=1,\ldots,n
  \label{eqn:dictionary_learning}
\end{align}
where $d_j$ denotes the $j$-th column of $D$.
This optimization problem is convex with respect to each of the
variables $\{s_i\}$ and $D$ when the other is fixed. Therefore, $s_i$
and $D$ can be alternatively updated recursively: solving the optimal sparse
coefficients $s_i$ given the dictionary $D$ fixed \myII{using Eq.~\eqref{eqn:lasso}}, and then learning
the dictionary $D$ based on the fixed sparse coefficients $s_i$
\cite{Lee+etal07:sparseCoding}. More efficient implementations via
online learning are also available~\cite{NIPS2007_3323,Mairal:2009:ODL:1553374.1553463}. We
here use the dictionary learning toolbox SPAMS from \cite{Mairal:2009:ODL:1553374.1553463} to solve
Eq.~\eqref{eqn:dictionary_learning}. SPAMS uses the LARS-Lasso
algorithm, which is a homotopy method providing the solutions for all
possible values of the regularization parameter
$\lambda$\cite{Efron_leastangle,Osborne01072000}. One of the advantages of
SPAMS is that it uses block-coordinate descent with warm restarts,
which guarantees convergence to a global optimum\cite{citeulike:1859441}.

\section{Adaptive Sampling and Reconstruction in Hyperspectral Imaging}\label{sec:methods}


A hyperspectral data cube consists of reflectance images at different
wavelengths. It consists of the $x$-$y$ spatial dimensions and the third
spectral dimension.

Our interest is in designing an efficient sampling and reconstruction
strategy for hyperspectral imaging based on
CS. CS for hyperspectral imaging relies on the existence of a
linear transformation $\mathcal T = \mathcal{S}_x \otimes
\mathcal{S}_y \otimes \mathcal{F}$ that gives a sparse representation
of the hyperspectral data cube, where $\mathcal{S}_x$,
$\mathcal{S}_y$, and $\mathcal{F}$ denote the linear transforms in
spatial domains and the spectral domain respectively, and $\otimes$
denotes the Kronecker product.

For the case of 2-dimensional spatial images,  it is well known that
variants of the DCT and wavelet transforms provide effective sparsifying
transformations. However, unlike the spatial domain, it turns out that neither the DCT nor
wavelet transform provides sufficient compressibility in the spectral
domain for
effective CS reconstruction (see~\autoref{fig:comparison} for example). We therefore turn our attention to
learning a sparsifying dictionary that uses \myII{training or
  historical data} of the
scene of interest. In fact, it has been observed that learning a
sparsifying dictionary from training datasets rather than using a
predetermined sparsifying basis (e.g. DCT or wavelets) usually gives
better sparse representation of the signal, and hence can improve the
CS reconstruction results\cite{1710377,Candes201159}. Moreover, it has
been discovered that the learned dictionary elements often
represent the basic spectral signatures comprising the
scene\cite{5762314,5898410}. Based on this, instead of using the random measurement matrices, we propose a
new adaptive sampling strategy using the \myII{existing} knowledge of the sparsifying
dictionary of the spectral signals. 

\subsection{Adaptive Sampling by SVD}
One of the important factors for successful recovery using CS is the
choice of the measurement
matrix. It is now well known that many types of random matrices whose
entries are drawn from Gaussian, Rademacher, or other sub-Gaussian
distributions can act as the measurement matrices for any signal that is
sparse with respect to an orthonormal basis. Researchers have also made substantial progress in the construction of sub-optimal deterministic universal measurement matrices to avoid the random
effects\cite{DeVore2007918,5208528,bourgain2011}. Such measurement
matrices are also used when the signal is sparse with respect to an overcomplete
dictionary. However, all these
constructions are for universal cases, which means they work for
arbitrary $k$-sparse signals. Less attention has been paid to
incorporating prior information, namely that a sparsifying dictionary
may be obtained in advance for the application in hand. It is natural
to ask the question whether we can sample smarter if the prior
knowledge, such as a dictionary learned from training data, is available.


We here consider this problem by applying SVD to the sparsifying
dictionary known in advance either from the analysis of historical
data or learned by dictionary learning. Then the measurement matrix can be
obtained by choosing the first $m$ singular vectors (associated with
the first $m$ singular values in descending order) of the sparsifying
dictionary.  Specifically, the economical SVD of the sparsifying
dictionary $D$ can be written as
\begin{align*}
  D = U \Sigma V^T,
\end{align*}
where $U\in \mathbb{R}^{d\times d}$ contains the left singular
vectors, $\Sigma \in \mathbb{R}^{d\times d}$ is a diagonal matrix
containing all the singular vectors, and $V\in\mathbb{R}^{n\times d}$
contains the right singular vectors. Now let $U_m$ denote the
submatrix of $U$ containing only the first $m$
columns of $U$. By setting the measurement matrix $\Phi = U_m^T$, we
obtain a deterministic sampling strategy. Then the sampling process is
simply a mixture of the signal of interest with fixed weights. 

The reconstruction problem can be further simplified due to this
special sampling strategy. Since
\begin{align*}
  U_m^T D
  &=
  U_m^T U \Sigma V^T \notag \\
  &=
  \Sigma_m V_m^T,
\end{align*}
where $\Sigma_m$ is a diagonal matrix containing the $m$ largest
singular values of the dictionary $D$ and $V_m$ contains the
corresponding  $m$ right singular vectors, we have from~\eqref{eqn:l1_noisy}
\begin{align}
  \min_s\|s\|_1 \quad \text{such that} \quad \|y - \Sigma_m V_m^T s\|_2<\epsilon,
\end{align}
which can be solved using the SPGL1 solver.


\subsection{Matrix Balancing}
A sufficient condition for CS to work is the Restricted
Isometry Property (RIP)\cite{1542412}, which requires that the singular
values of the sensing matrix do not vary too much. In other words, the
sensing matrix needs to be well
conditioned. We therefore further condition the sensing
matrix by applying matrix balancing. Matrix balancing has been widely
studied in various disciplines for square matrices
\cite{Chen2000261,Knight26102012}. A square matrix is said to be
balanced with respect to the $\alpha$-norm if for any index $i$, the $\alpha$-norm of the $i$th row
is the same as that of the $i$th column. We here introduce a new
balancing algorithm for general rectangular matrices based on SVD. We
say an $m$ by $n$ ($m<n$) rectangular matrix is balanced if the row
norms are the same as the column norms multiplied by a factor of
$n/m$. The details of this algorithm is shown in
Algorithm~\ref{alg:balance}. It decomposes a given matrix $A$ into $A =
PBQ$, where $P$ is a square invertible matrix, $Q$ is a nonsingular
diagonal matrix, and $B$ is balanced in the sense that its rows are
orthonormal and
column norms only differ from the row norms by a constant factor
provided the algorithm converges.

\begin{algorithm}
  \caption{Matrix Balancing by SVD}
  \begin{algorithmic}[1]
    \STATE\textbf{Input:} matrix $A$, maximum number of iterations $t_{\max}$.
    \STATE\textbf{Initialization:} Set $t = 1$, $B_1 = A$, and $P_1$
    and $Q_1$ to be identity matrices.
    \WHILE {$t < t_{\max}$}
      \STATE Perform SVD on $B_t$ $$B_t = U_t \Sigma_t V_t^T = U_t \Sigma_t
      (V_t^T S_t) S_t^{-1},$$ where $S_t$ is a diagonal scaling matrix
      to normalize columns of $V_t^T$.
      \STATE Update: $$P_{t+1} = P_tU_t\Sigma_t,$$ $$B_{t+1} =
      V_t^TS_t,$$ $$Q_{t+1} = S_t^{-1} Q_t.$$
    \ENDWHILE
    \STATE \textbf{Output:} matrix $A$ is decomposed into $A =
    P_{t_{\max}} B_{t_{\max}} Q_{t_{\max}}$, where $P_{t_{\max}}$ is
    nonsingular, $Q_{t_{\max}}$ is diagonal and nonsingular,
    $B_{t_{\max}}$ is balanced.
  \end{algorithmic}
  \label{alg:balance}
\end{algorithm}

Now let $\tilde{y} = P^{-1}y$ and $\tilde{s} = Qs$. Then the minimization problem~\eqref{eqn:l0} can be
rewritten as
\begin{align}\label{eqn:l0_weighted}
  \min_{\tilde{s}} \|Q^{-1}\tilde{s}\|_0 \quad \text{subject to}\quad \tilde{y}=B\tilde{s}.
\end{align}
 It is clear that $\tilde{s}$ has the same sparsity as $s$ since $Q$
 is diagonal and nonsingular. Thus, problem~\eqref{eqn:l0_weighted} is
 equivalent to
 \begin{align*}
  \min_{\tilde{s}} \|\tilde{s}\|_0 \quad \text{subject to}\quad \tilde{y}=B\tilde{s},
\end{align*}
which is equivalent to the $\ell_1$ minimization problem
 \begin{align*}
  \min_{\tilde{s}} \|\tilde{s}\|_1 \quad \text{subject to}\quad \tilde{y}=B\tilde{s}.
\end{align*}
Similarly, it can also be written as a BPDN problem in presence
of noise
\begin{align}\label{eqn:l1_conditioned}
  \min_{\tilde{s}}\|\tilde{s}\|_1 \quad \text{subject to}\quad
  \|\tilde{y} - B\tilde{s}\|_2<\epsilon.
\end{align}
Moreover, the sensing matrix is now better conditioned
since it is normalized with respect to both rows and columns, and
hence may improve the RIP constant since it is directly associated with
the largest and smallest singular values of the submatrices. The BPDN
problem~\eqref{eqn:l1_conditioned} can be easily solved by using the
SPGL1 solver.

\myII{In summary, our sampling and reconstruction procedure can be
  described as follows. We first find a sparsifying dictionary $D$ of the
  spectral signals of interest using dictionary learning based on
  training or historical data. We then construct a CS measurement
  matrix $U_m$ by applying SVD to the learned dictionary, using only
  the first $m$ significant singular vectors. Applying $U_m$ to the
  signal of interest yields a measurement vector of length $m$, which
  is less than the length of the signal of interest $n$. Now by feeding
the measurement vector and the sensing matrix $A = U_mD$ into the
standard CS solvers, we are able to obtain the signal of interest with
length $n$.}

\section{\myII{Numerical} Experiments}
\label{sec:evaluation}
\myII{As discussed in Section~\ref{sec:methods}, the sparsifying
transformations for 2-dimensional spatial images are well known. One
can thus easily apply conventional CS methods in the spatial domain
for efficient sampling. We now describe the results of numerical
experiments that demonstrate the advantage of using the combination of
dictionary learning and SVD over other combinations for the spectral
domain signals.}

\begin{figure*}[ht!]
  \centering
  \begin{subfigure}{0.34\textwidth}
    \includegraphics[width = \textwidth]{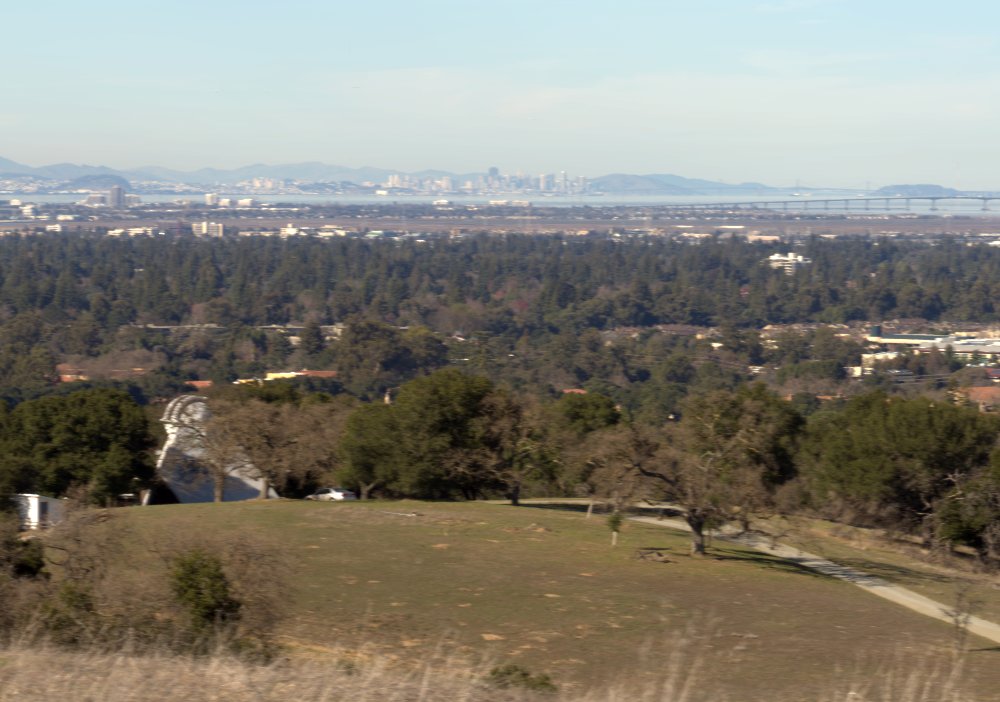}
    \caption{San Francisco}
    \label{fig:sf}
  \end{subfigure}
  \begin{subfigure}{0.3\textwidth}
    \includegraphics[width = \textwidth]{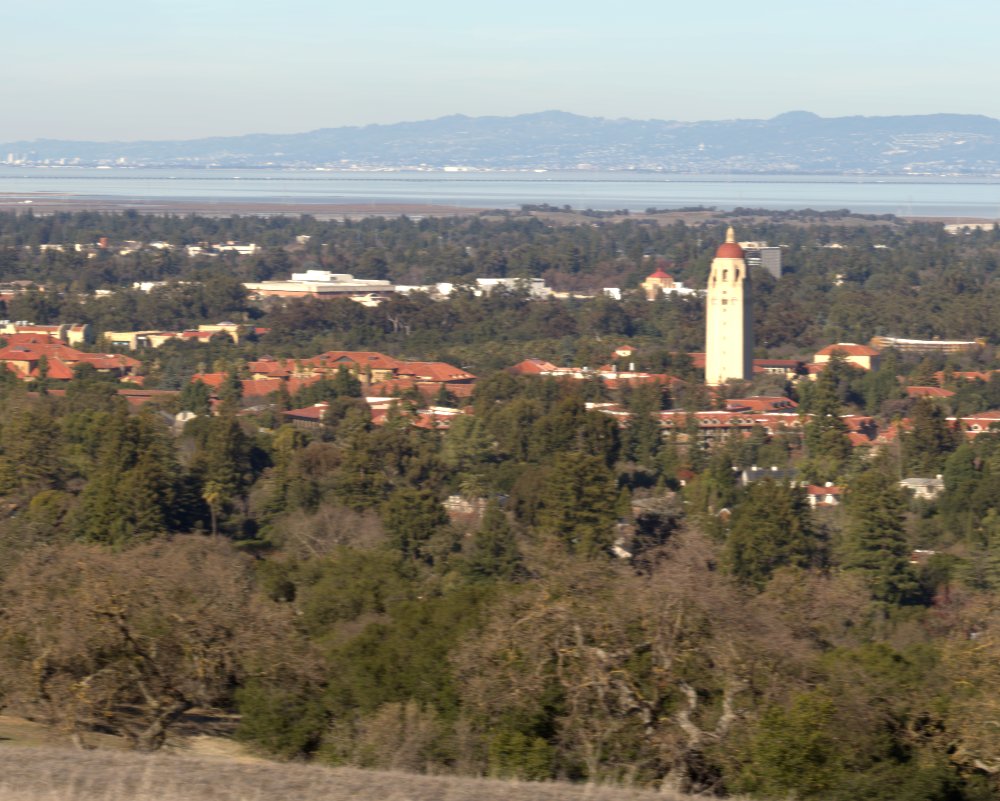}
    \caption{Stanford Tower}
    \label{fig:st}
  \end{subfigure}
  \begin{subfigure}{0.24\textwidth}
    \includegraphics[width = \textwidth]{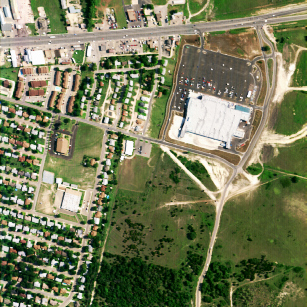}
    \caption{Urban}
    \label{fig:urban}
  \end{subfigure}
  \caption{RGB images of the three hyperspectral datasets used.}
  \label{fig:datasets}
\end{figure*}

\subsection{Dataset}


In the following numerical experiments we will be using three hyperspectral datasets. The
first two are both from the SCIEN lab at Stanford
University~\cite{Skauli13}. The landscape hyperspectral image of
San Francisco (SF) has a spatial resolution of 1000 $\times$ 702, and
the hyperspectral image of the Stanford Tower has a
spatial resolution of 1000 $\times$ 801, both containing 148 spectral
bands ranging from 415nm to 950nm. The third dataset is the Urban
dataset from the US Army Corps of Engineers, which is available at
\url{www.agc.army.mil/Missions/Hypercube.aspx}. It has spatial
resolution of 307 $\times$ 307, and contains 210 spectral bands
ranging from 399nm to
2499nm. ~\autoref{fig:datasets} provides a view of the three datasets as RGB images.

\begin{figure}[ht!]
  \centering
    \begin{subfigure}{0.44\textwidth}
    \includegraphics[width=\textwidth]{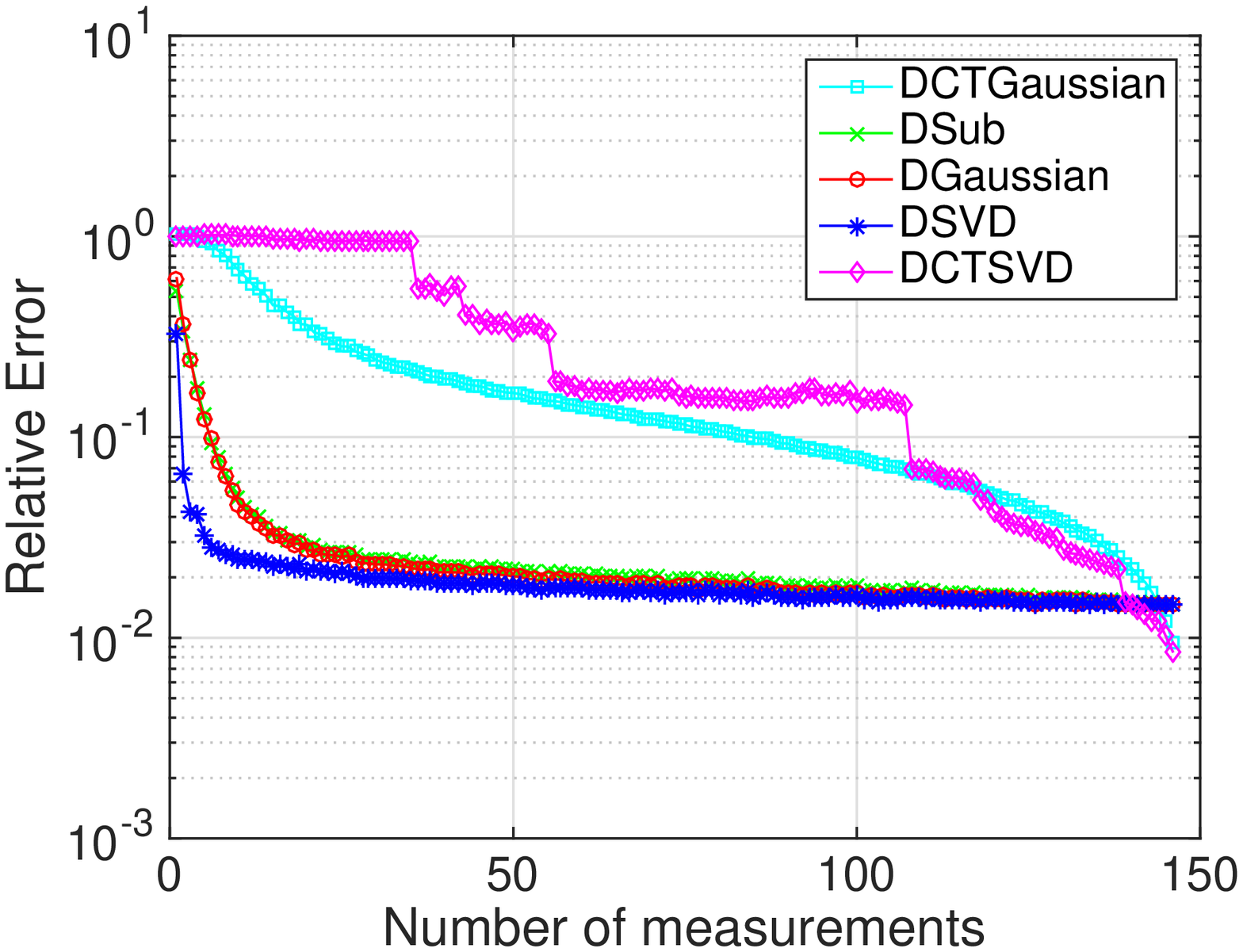}
    \caption{San Francisco}
    \label{fig:comparison_SF}
  \end{subfigure}
    \begin{subfigure}{0.44\textwidth}
    \includegraphics[width=\textwidth]{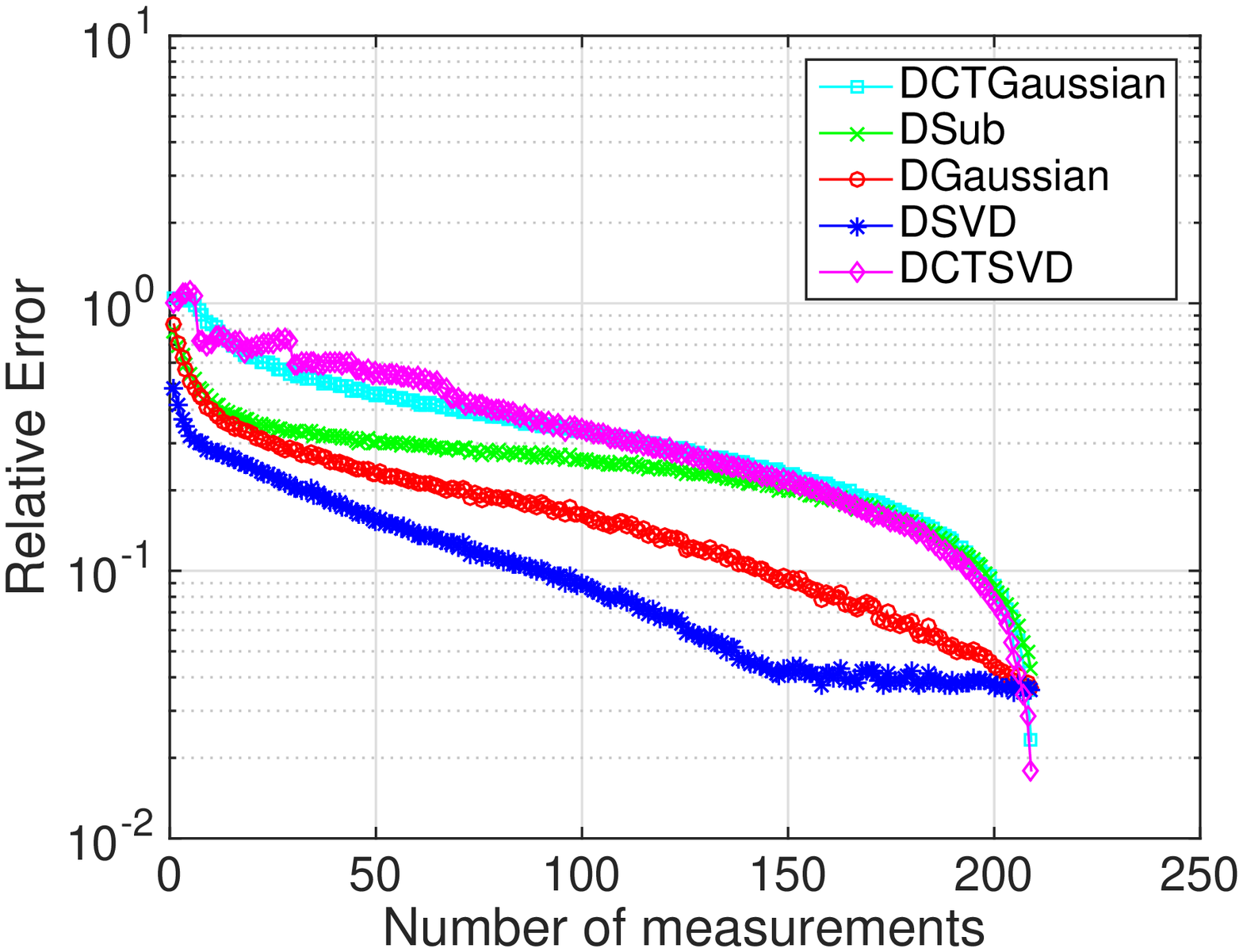}
    \caption{Urban}
    \label{fig:comparison_urban}
  \end{subfigure}
  \caption{Comparison among different sampling and sparsifying
    strategy combinations}
  \label{fig:comparison}
\end{figure}

\subsection{Comparisons of sampling and sparsifying strategies}

\begin{figure*}[hb!]
  \centering
  \begin{subfigure}{0.3\textwidth}
    \includegraphics[width = \textwidth]{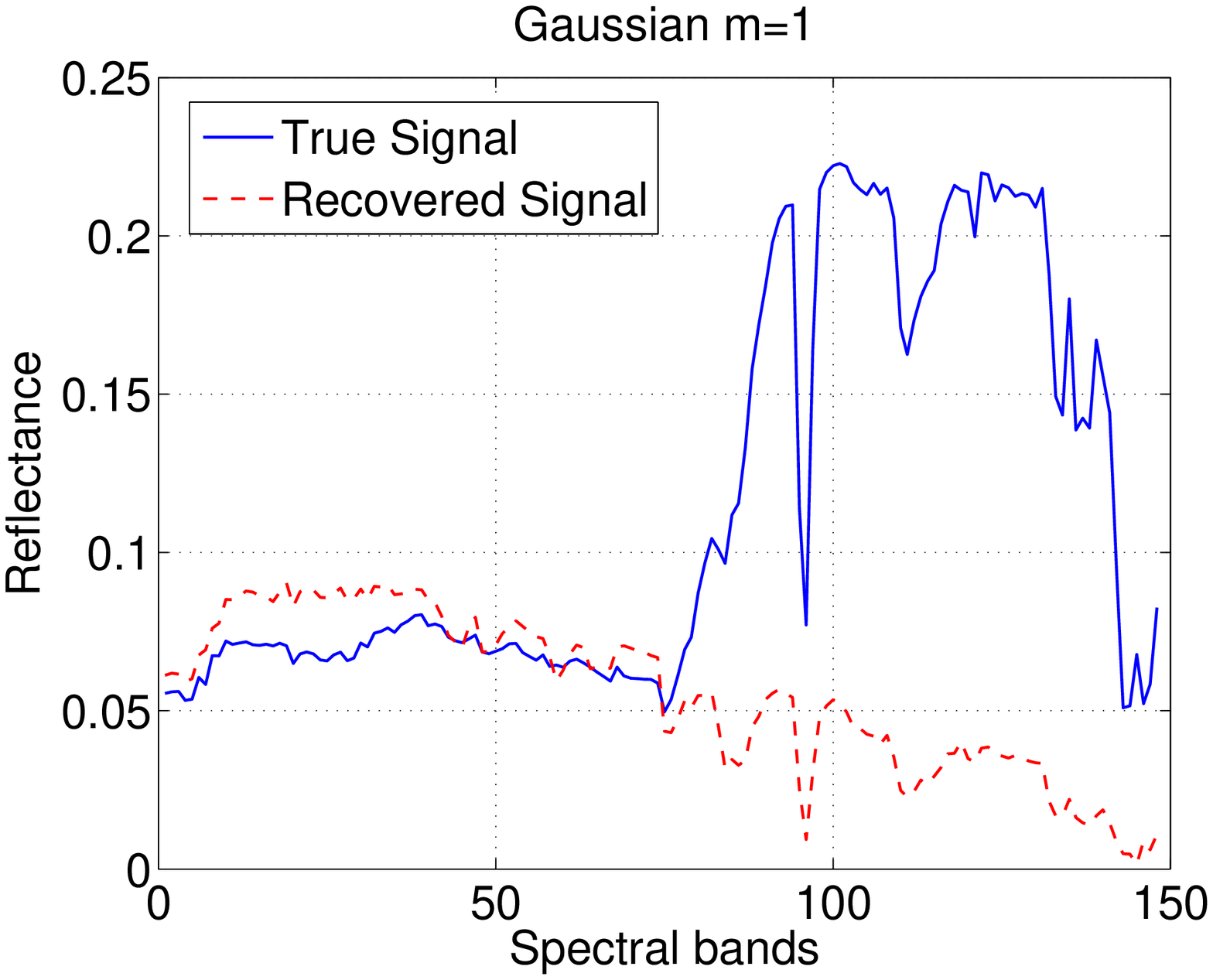}
    \caption{}
  \end{subfigure}
  \begin{subfigure}{0.3\textwidth}
    \includegraphics[width = \textwidth]{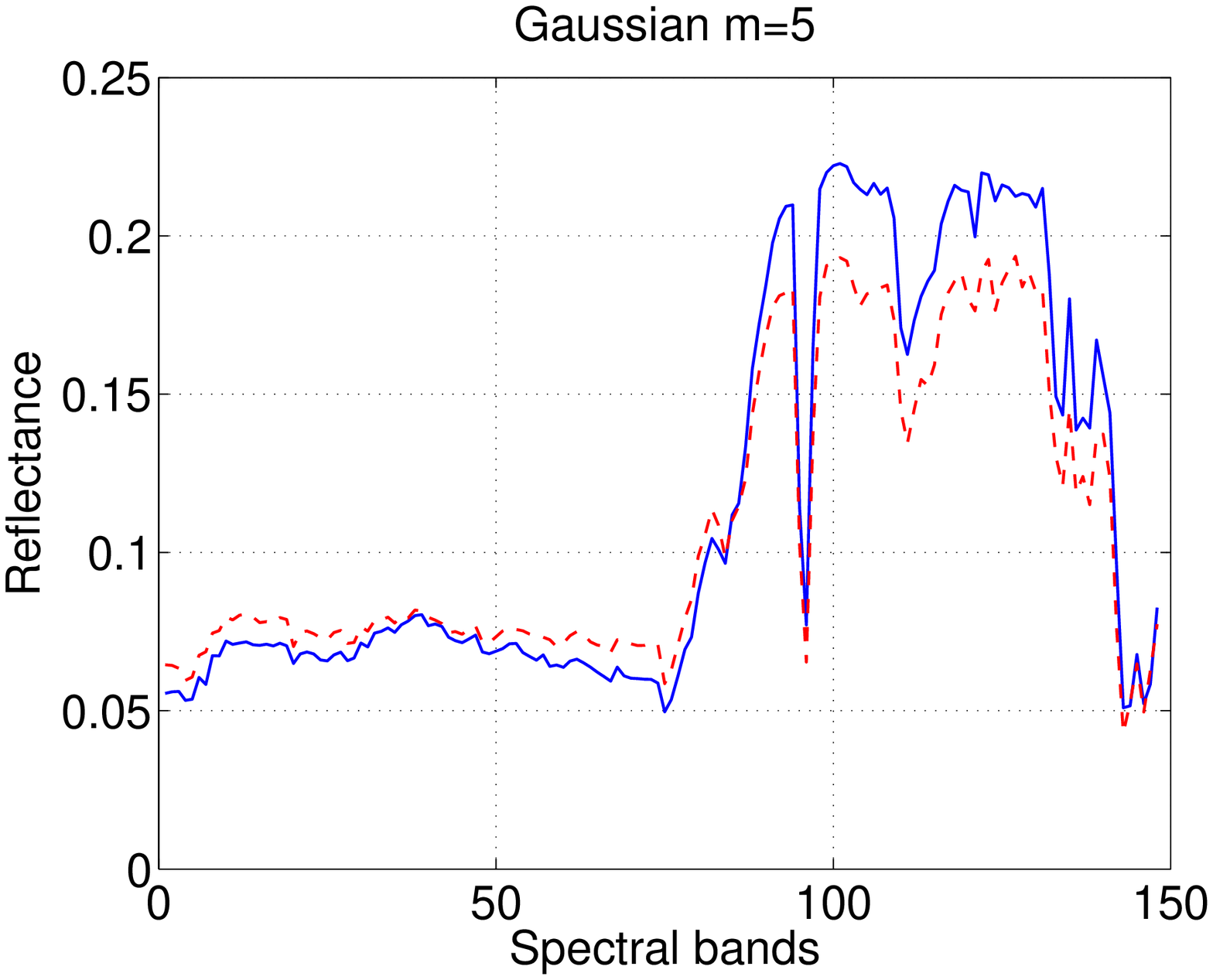}
    \caption{}
  \end{subfigure}
  \begin{subfigure}{0.3\textwidth}
    \includegraphics[width = \textwidth]{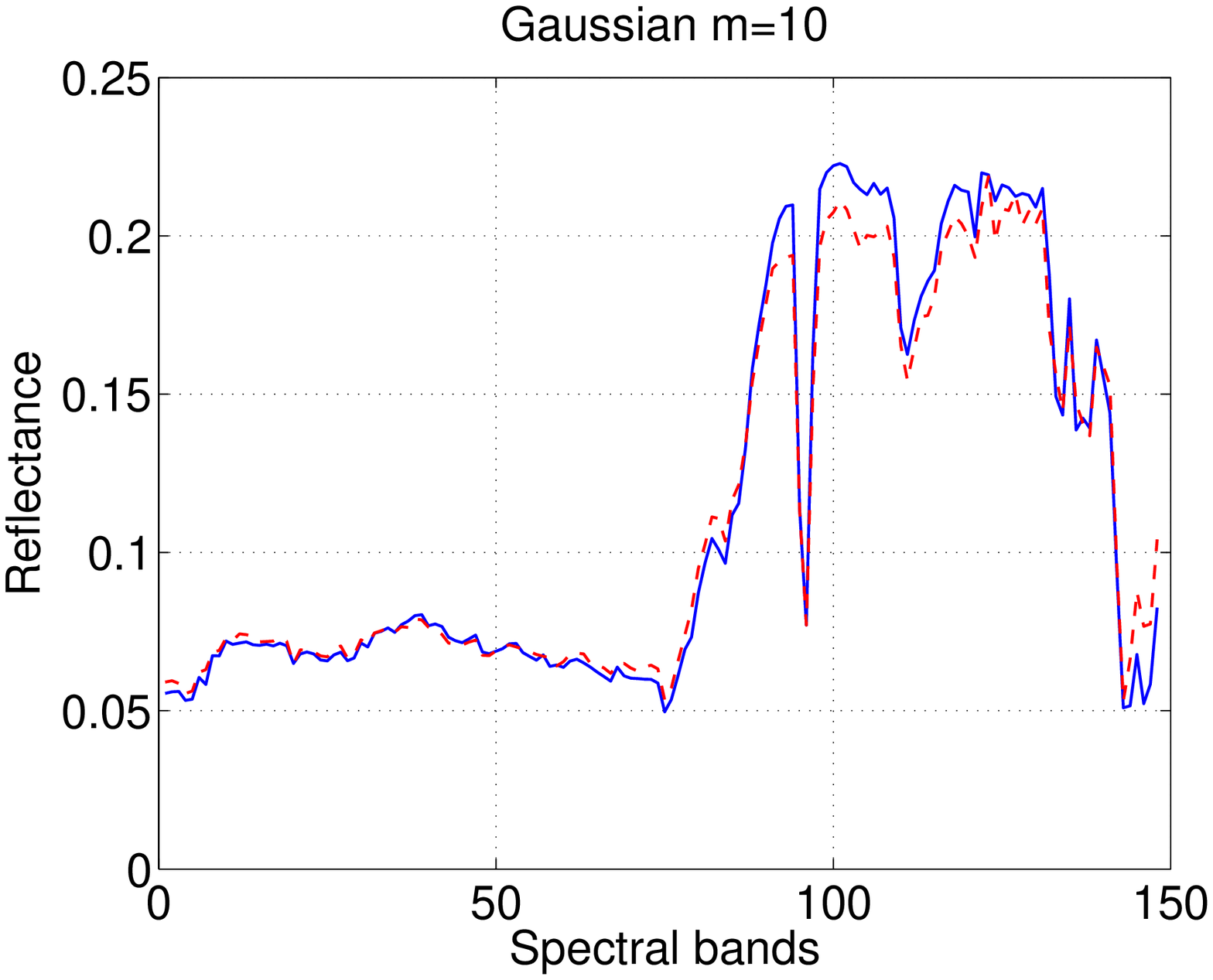}
    \caption{}
  \end{subfigure}
  \begin{subfigure}{0.3\textwidth}
    \includegraphics[width = \textwidth]{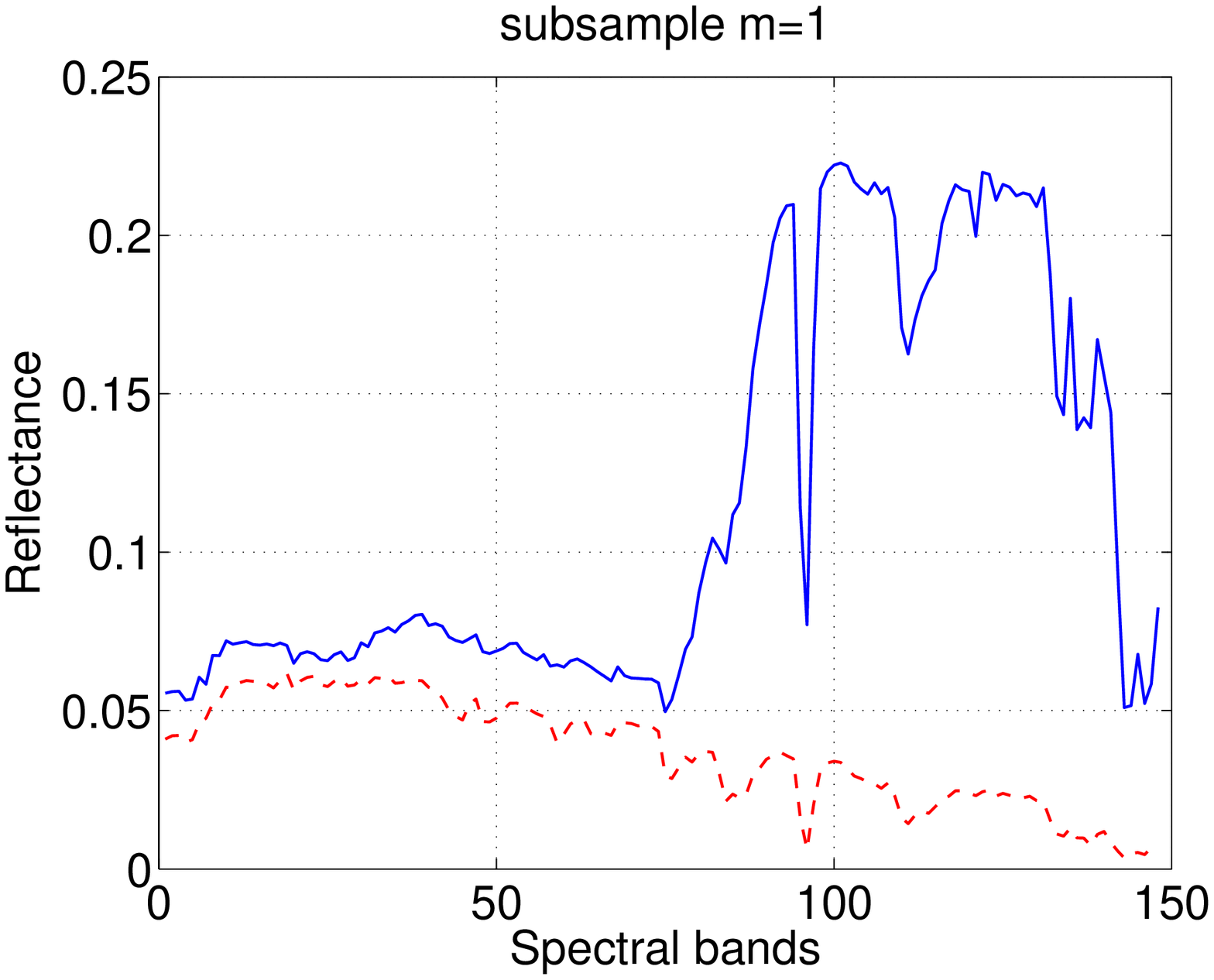}
    \caption{}
  \end{subfigure}
  \begin{subfigure}{0.3\textwidth}
    \includegraphics[width = \textwidth]{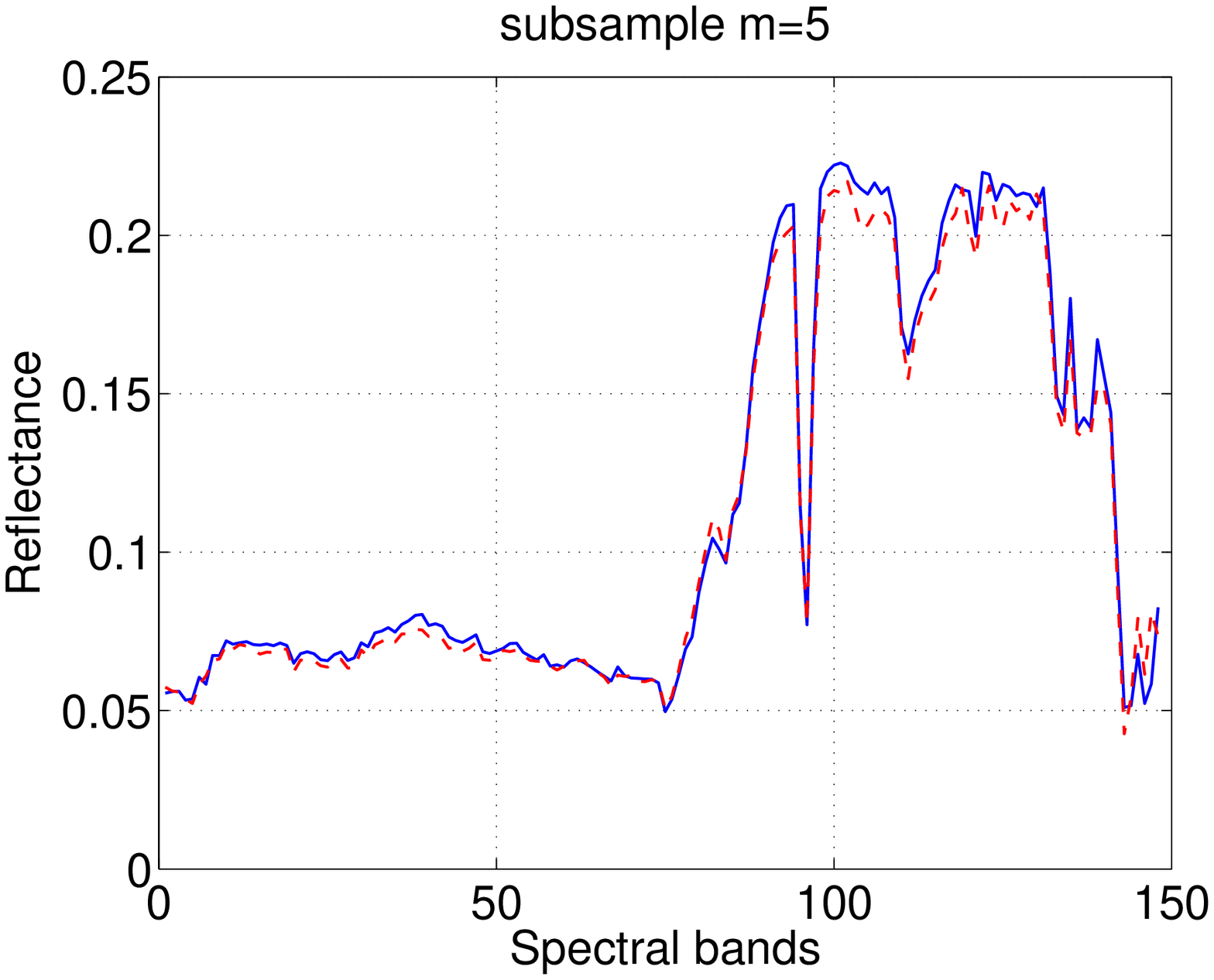}
    \caption{}
  \end{subfigure}
  \begin{subfigure}{0.3\textwidth}
    \includegraphics[width = \textwidth]{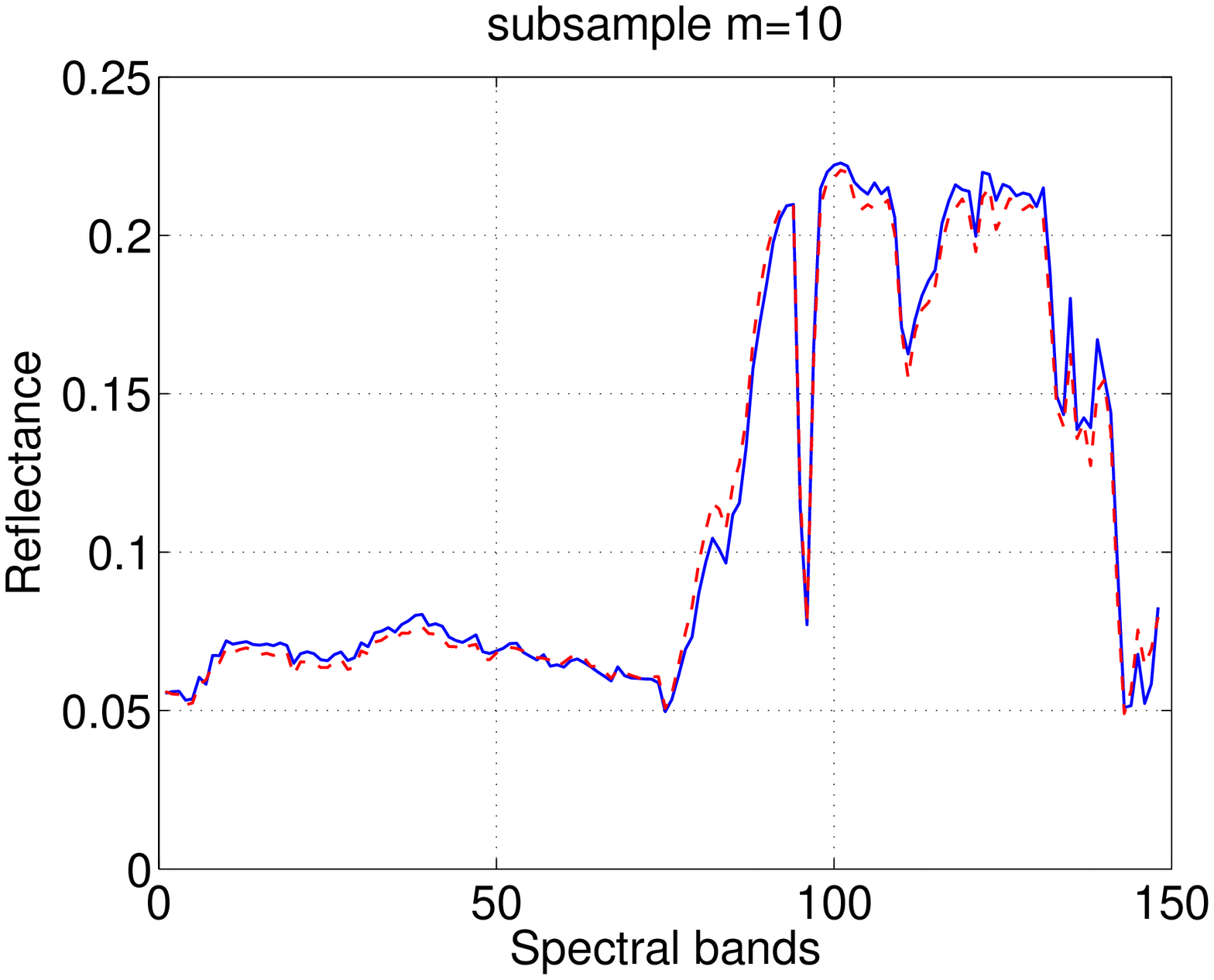}
    \caption{}
  \end{subfigure}
  \begin{subfigure}{0.3\textwidth}
    \includegraphics[width = \textwidth]{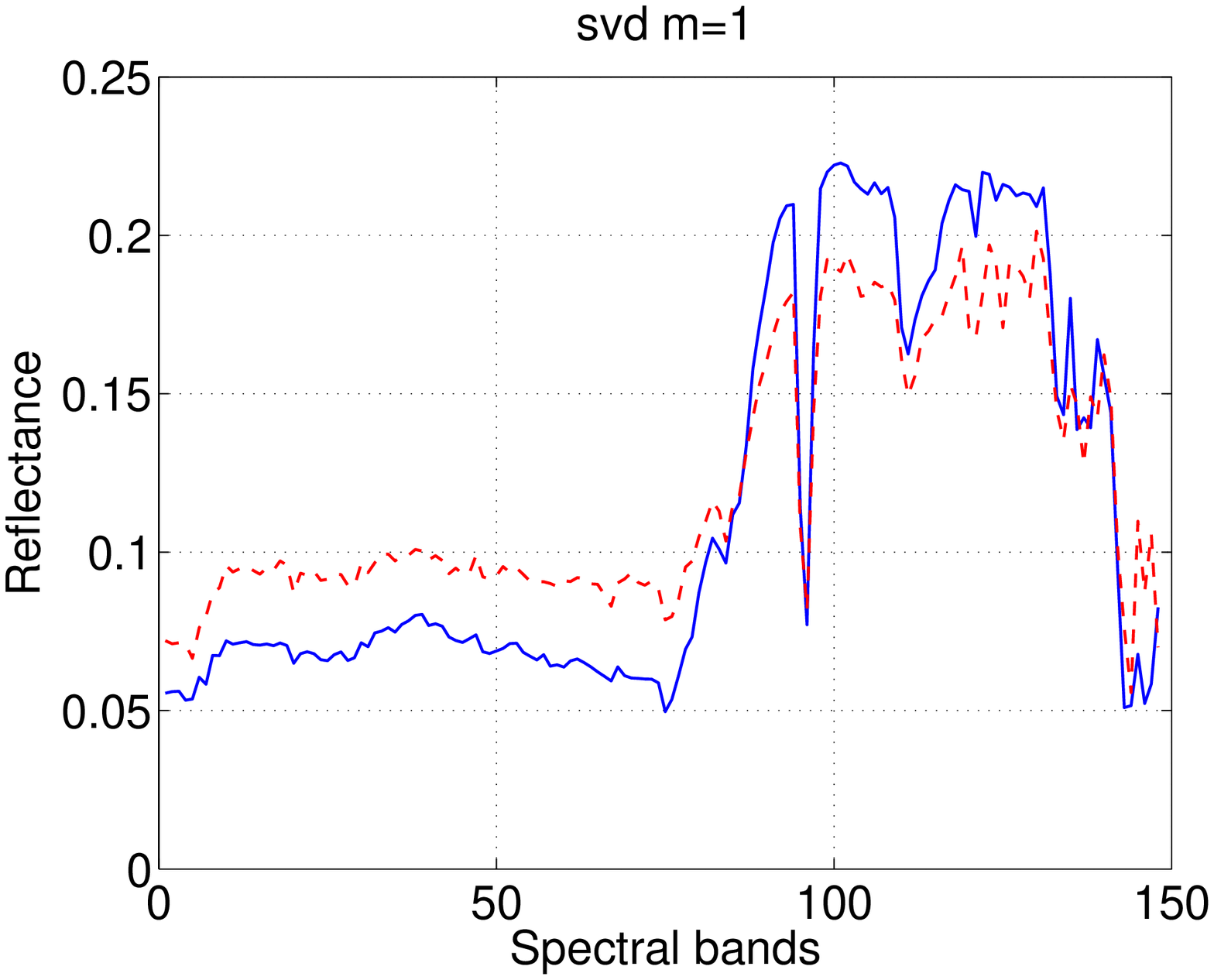}
    \caption{}
  \end{subfigure}
  \begin{subfigure}{0.3\textwidth}
    \includegraphics[width = \textwidth]{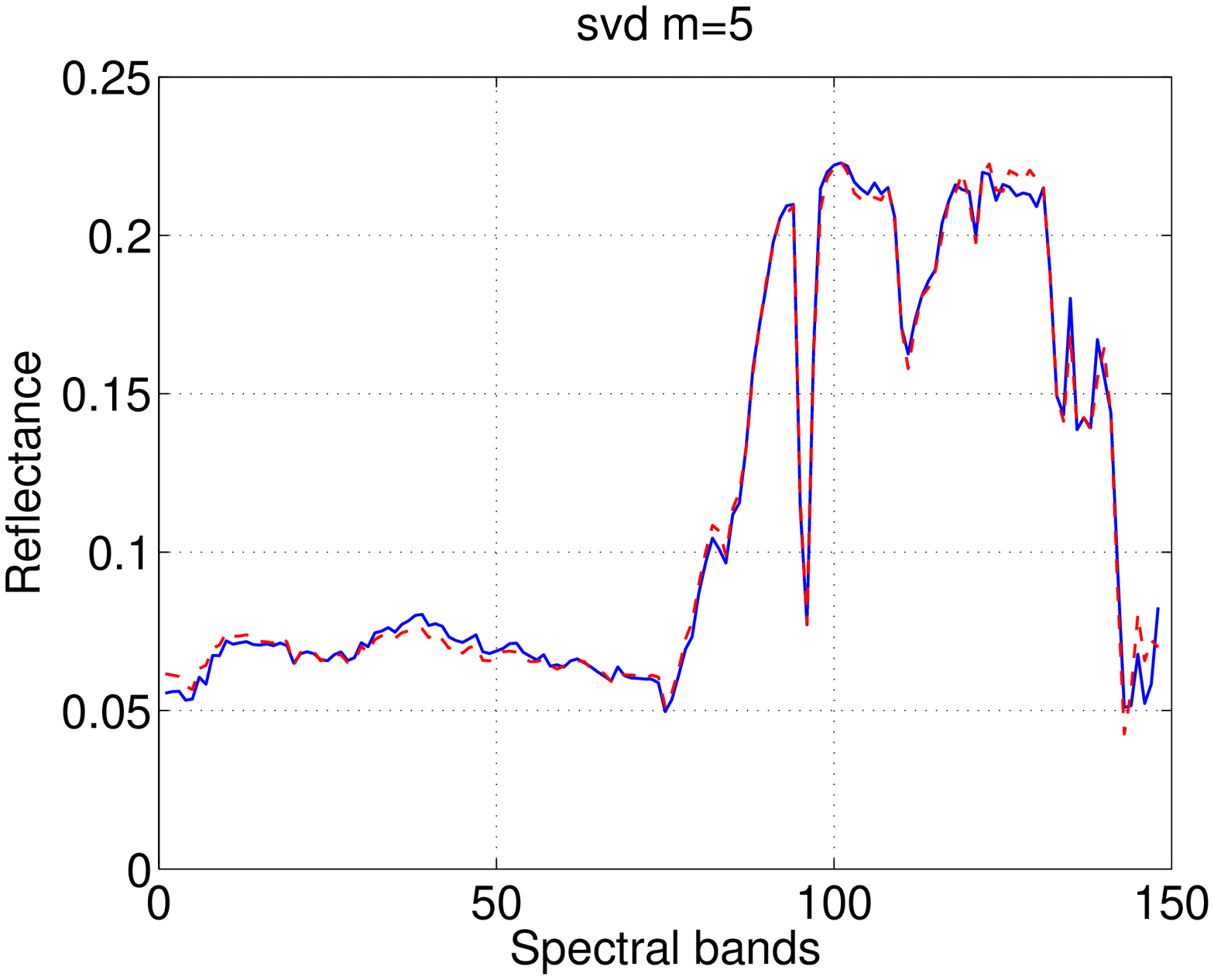}
    \caption{}
  \end{subfigure}
  \begin{subfigure}{0.3\textwidth}
    \includegraphics[width = \textwidth]{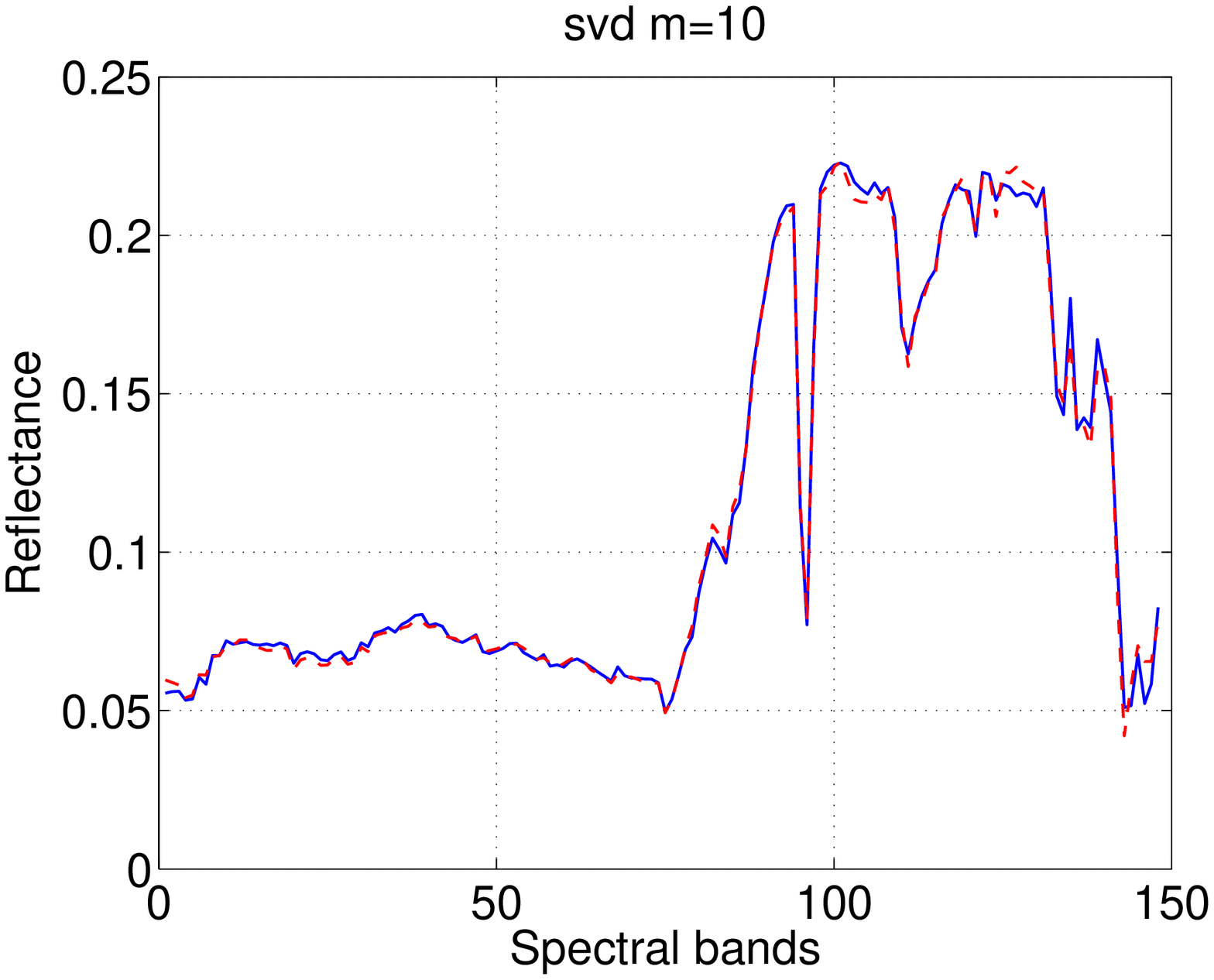}
    \caption{}
  \end{subfigure}
  \caption{Comparison of the performance of different dictionary and
    measurement strategies combinations for a randomly chosen pixel in
    the San Francisco
    dataset. Figure (a)-(c)
    describe the reconstruction performance of DGaussian with
    different number of measurements $m$; figure (d)-(f)
    describe the reconstruction performance of DSub with
    different number of measurements $m$; figure (g)-(i)
    describe the reconstruction performance of DSVD with
    different number of measurements $m$.}
  \label{fig:comparison2}
\end{figure*}

The sparsifying dictionaries we use for our experiment are learned
from the training data, which are obtained by randomly choosing half
of the spectral signals (half of the pixels in the spatial
domain) from the hyperspectral datasets mentioned above. The rest of the datasets are kept as the test data. \myII{We have
normalized the hyperspectral data for each pixel to unit $\ell_2$ norm.
We choose the error bound $\epsilon$ for BPDN in Eq.~\eqref{eqn:l1_noisy} to be
$0.01$ and use the regularization parameter $\lambda=1.2/\sqrt{d}$ for
sparse coding in Eq.~\eqref{eqn:dictionary_learning}, as suggested
by~\cite{Mairal:2009:ODL:1553374.1553463}. Small variations of these
parameters have yielded very similar performance.} We now compare the performance of the five different combinations of
sparsifying domains and sensing methodology using the SPGL1
solver. \autoref{fig:comparison_SF} and~\autoref{fig:comparison_urban}
compare their performance for the landscape
hyperspectral image of San Francisco and the remote sensing
hyperspectral image of an urban area respectively. Here DCTGaussian
uses a DCT sparsifying matrix and a Gaussian measurement
matrix. DSub means sparsifying matrix from dictionary learning and
uniformly random subsampling. DGaussian stands for sparsifying matrix
from dictionary learning with a Gaussian
measurement matrix. DSVD stands for sparsifying matrix from dictionary
learning with measurement matrix from SVD. \myII{We also include
  another combination DCTSVD, using DCT for the sparsifying matrix and
  SVD for the measurement matrix, for comparison.} The
relative error here is defined as
\begin{align*}
  \text{Relative Error }=\frac{\|x^*-x\|_2}{\|x\|_2},
\end{align*}
where $x^*$ is the reconstructed signal.
It is clear from the plots that the
conventional combination DCTGaussian does not provide efficient
reconstruction, and performs much
worse than the other combinations. In fact, it
performs even worse than DSub, a naive choice of the measurement
matrix combined with dictionary learning. \myII{The DCTSVD results are
not good either as expected, since the DCT basis is a general
sparsifying basis, which does not contain any information from the
data. In addition, its singular values are all ones. Therefore, performing SVD on
the DCT basis will not provide additional benefits.} For the other three
combinations, the
combination DSub performs almost the same as 
DGaussian on the SF dataset, and worse on the Urban dataset as expected. The combination DSVD, as
described in Section~\ref{sec:methods}, performs significantly better than the
others on both datasets. \myII{One may notice that the performance for all
the methods on the Urban data is worse than that on the SF
data. This in fact coincides with our intuition. First of all, the
number of measurements needed is directly related to the number of
underlying spectral features mixed together in each pixel. In other
words, it is related to how sparse the spectrum of the pixel is in
terms of the learned dictionary. The SF data has a larger spatial resolution and
relatively simpler spectral features, resulting in less mixtures for
each pixel. While the Urban data has a smaller spatial resolution and
much more complicated features, which could result in more feature mixtures
in each pixel, and thus require larger number of measurements for good
reconstruction.} 

We next give a more straightforward view of the reconstruction
performance of the three combinations mentioned above using dictionary
learning as the sparsifying domain, namely DSub, DGaussian, and DSVD. \autoref{fig:comparison2} uses the same dataset as in~\autoref{fig:comparison_SF}, namely the SF dataset. All the figures describe the
spectrum of the same randomly chosen pixel \myII{(representing woods in this case)}. We choose the number
of measurements $m$ to be 1, 5, and 10 respectively. The blue curve is the
ground truth of the spectrum for this pixel, while the red dashed curve
is the reconstructed spectrum. As one can see, the Gaussian sampling and
uniformly random subsampling strategies have about the same performance. They both fail when the
number of measurement is 1 (0.68\%) and start to recover the spectrum
better with more measurements. The combination of dictionary learning
with SVD performs
the best in the sense that even when the number of measurement is 1, the
reconstruction has already been able to catch a rough shape of the true
spectrum and, when the number of measurement is 5 (3.38\%), it can
recover the true spectrum accurately. 

The above results indicate that this approach can bring significant savings in cost,
processing time, and storage. 
The amount of data
generated by the conventional 2-D hyperspectral cameras causes
difficulty for local storage or transmission. In contrast, with DSVD
\myII{applied to the spectral domain}, it is possible to store/transfer less
than 5\% of the original amount of data, which could be handled
efficiently. For instance, the amount of data for a single
$1000\times1000\times600$ hyperspectral cube to be stored/transmitted using
conventional hyperspectral imager could easily reach 600 megabytes,
while using the proposed strategy, it is less than 30
megabytes. \myII{Moreover, one may apply the conventional CS method,
  or even common image compression methods (e.g. JPEG) to the spatial
  images using the tensor structure as discussed in
  Section~\ref{sec:methods} to further improve the efficiency.}

\subsection{Matrix Balancing}

\begin{figure*}[ht!]
  \centering
  \begin{subfigure}{0.44\textwidth}
    \includegraphics[width=\textwidth]{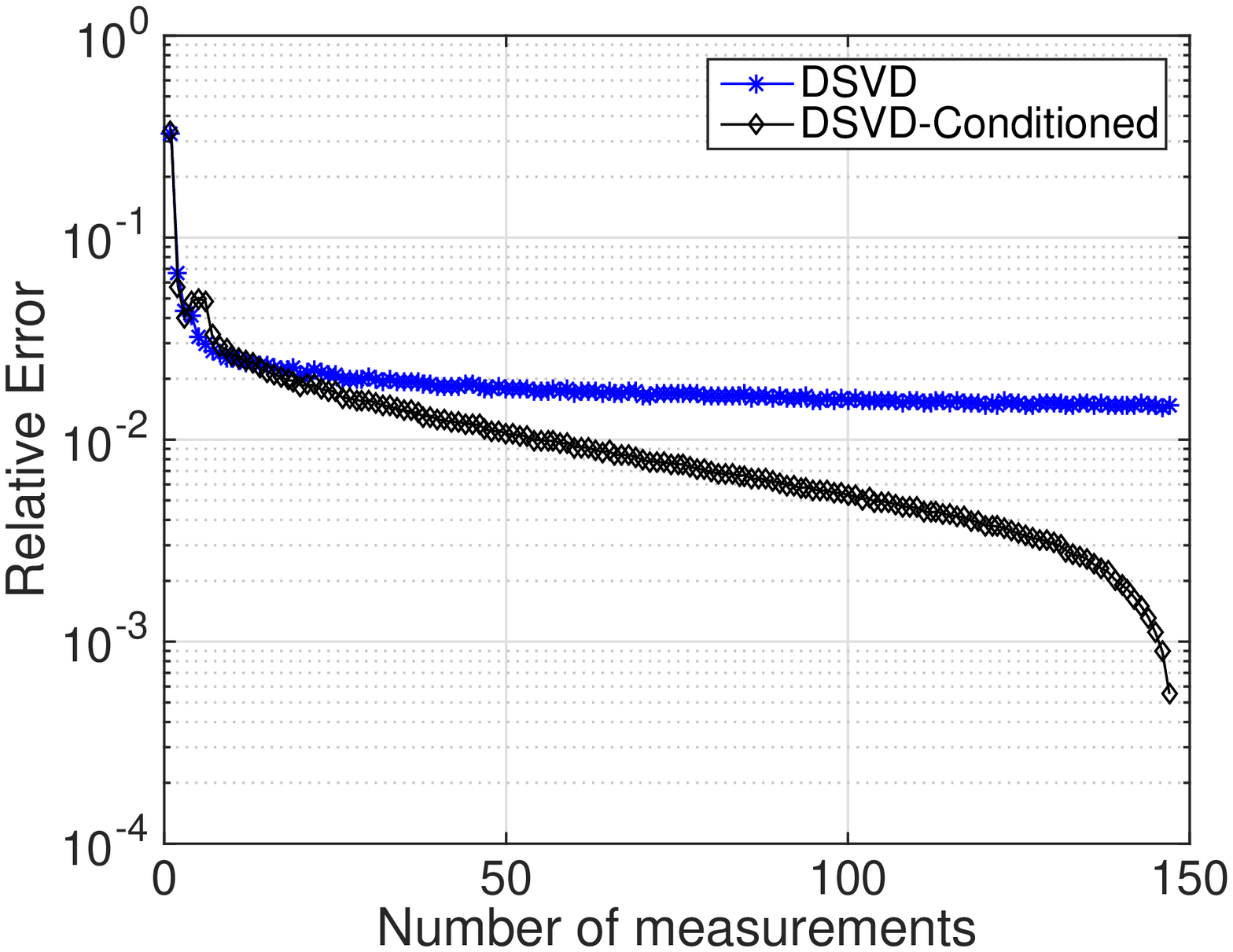}
    \caption{San Francisco}
    \label{fig:balancing_SF}
  \end{subfigure}
  \begin{subfigure}{0.44\textwidth}
    \includegraphics[width=\textwidth]{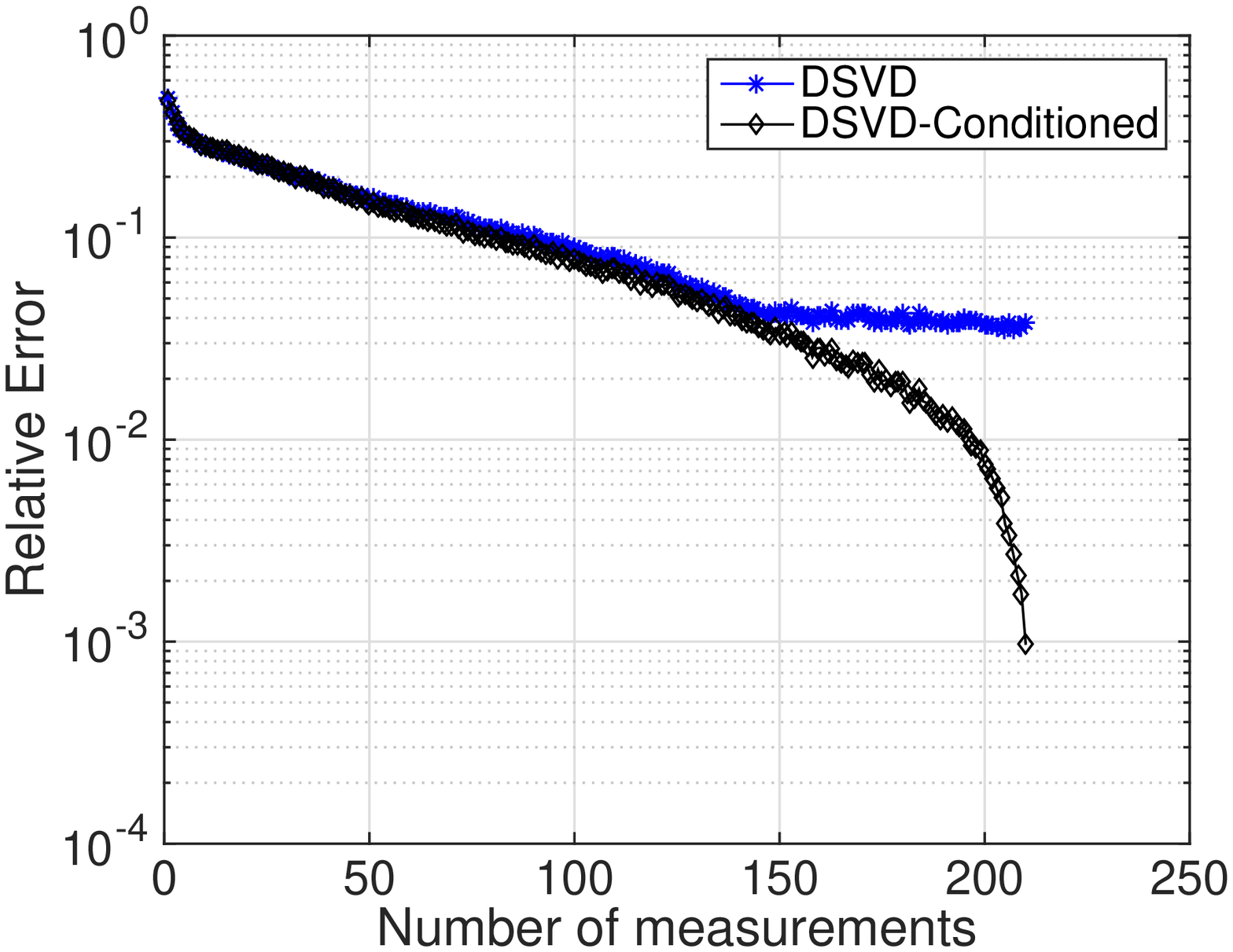}
    \caption{Urban}
    \label{fig:balancing_Urban}
  \end{subfigure}
  \caption{Effect of sensing matrix balancing on reconstruction
    performance}
  \label{fig:balancing}
\end{figure*}

Next we demonstrate that matrix balancing provides further
enhancement in the reconstruction performance. 
The results of performing matrix balancing on the
sensing matrix as described in Algorithm~\ref{alg:balance} \myII{for 10
iterations} are shown
in~\autoref{fig:balancing}.
From the figures we see clearly that
matrix balancing does improve the reconstruction performance. However,
the benefit is more significant on the SF dataset as shown
in~\autoref{fig:balancing_SF} than on the Urban dataset shown
in~\autoref{fig:balancing_Urban}.
\begin{figure*}[ht!]
  \centering
  \begin{subfigure}{0.44\textwidth}
    \includegraphics[width=\textwidth]{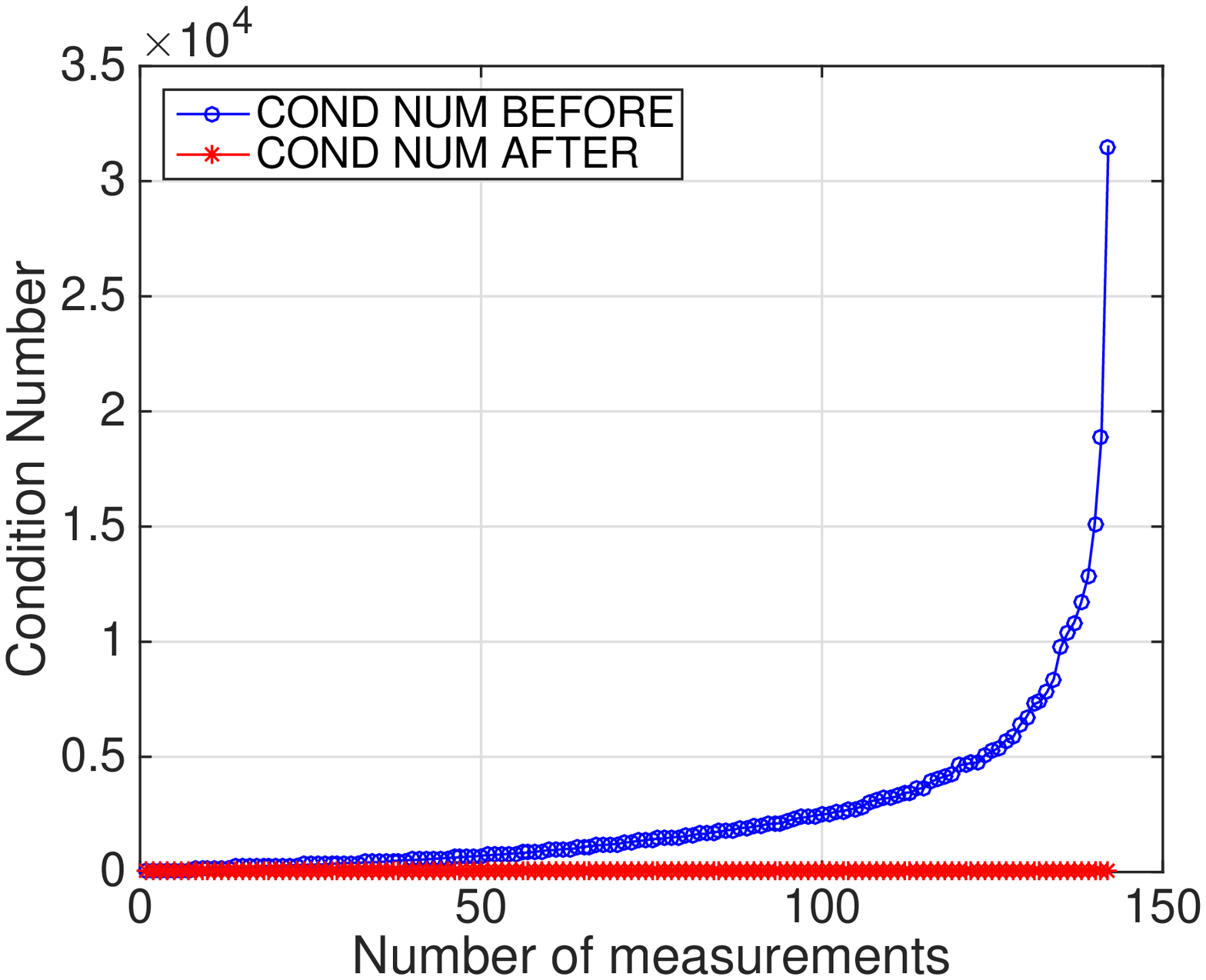}
    \caption{San Francisco}
    \label{fig:cond_num_SF}
  \end{subfigure}
  \begin{subfigure}{0.44\textwidth}
    \includegraphics[width=\textwidth]{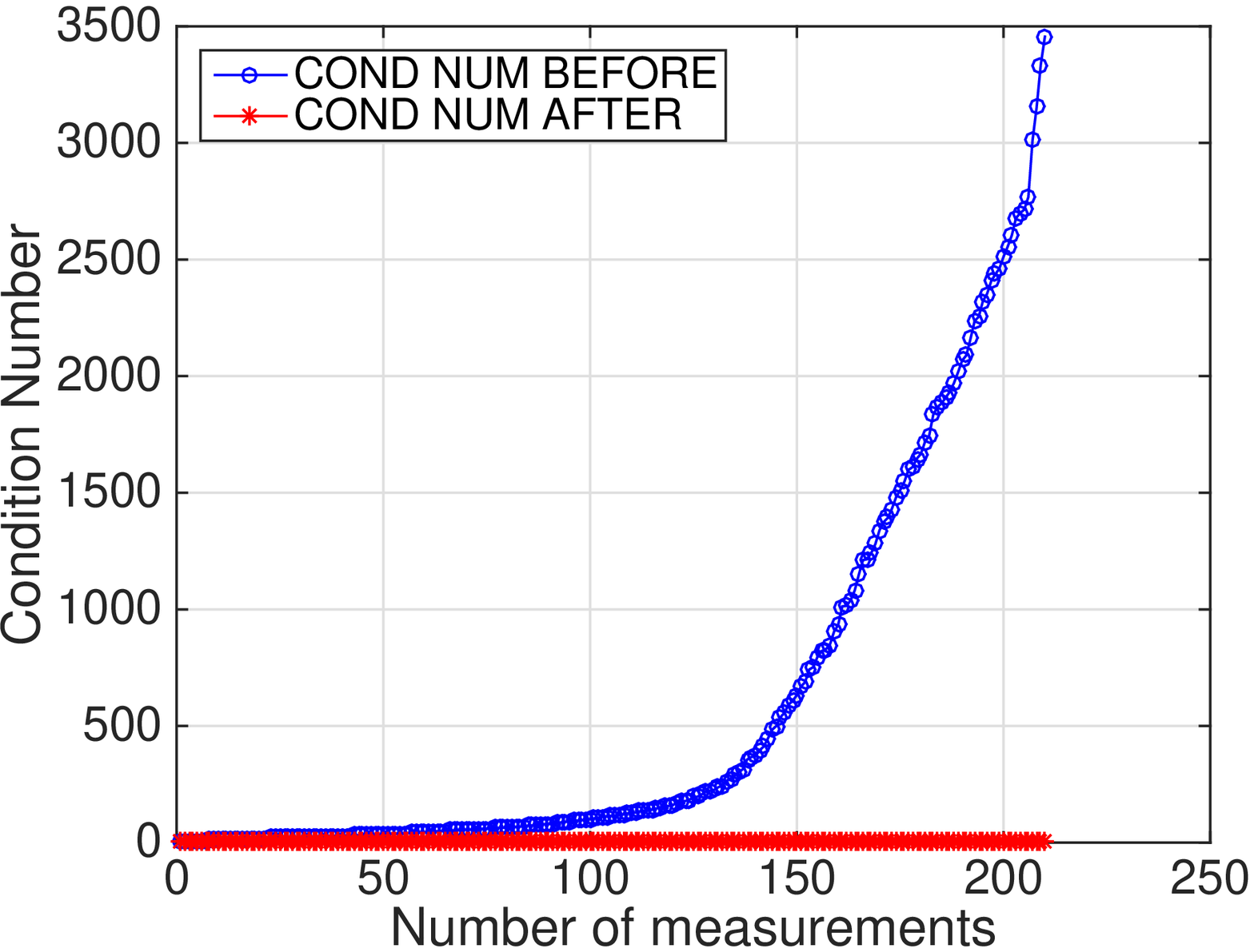}
    \caption{Urban}
    \label{fig:cond_num_Urban}
  \end{subfigure}
  \caption{Condition number of the sensing matrix before and after
    matrix balancing}
  \label{cond_num}
\end{figure*}
To understand this phenomenon, we investigate the changes in condition
numbers of the sensing matrices. Specifically, we check for both
datasets the condition numbers of the sensing matrices with the number
of measurements varying before and after performing matrix
balancing. \autoref{cond_num} plots the behaviors of the condition
numbers of the sensing matrices for both datasets. From the figures, the condition numbers for the sensing
matrices for both datasets blow up as the number of measurements
increases. Looking more closely though, the condition numbers of the
sensing matrices for the SF dataset blow up significantly faster than
that for the Urban data as the number of measurements approaches the
full dimension, resulting in ill conditioned systems. Therefore, the
reconstruction performance gain for the SF dataset is much better than
that for the Urban dataset using matrix balancing as shown in~\autoref{fig:balancing}.









\my{
\subsection{Robustness}
The robustness of the dictionary learned from the training data has
previously been studied in \cite{5762314}. In that paper, authors learned the
dictionary from the Smith Island training data and tested it on the test data
collected on the same date. They showed by evaluation that the
reconstruction relative MSE was less than 0.09\% on that
dataset. In addition, they also tested the dictionary learned on a test
set collected in a different season for the same scene. They showed that the
reconstruction relative MSE in this scenario was less than 0.71\%,
which was worse than for the test data collected on the same day but was
still very good overall considering the vegetation and atmospheric
characteristics were highly likely to be different due to different
seasons.  

\begin{figure}[ht!]
  \centering
  \includegraphics[scale=0.4]{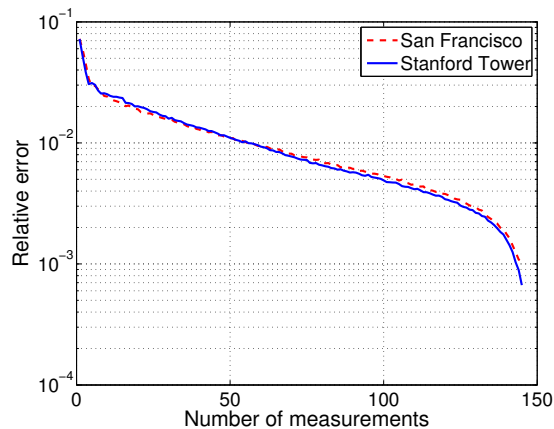}
  \caption{Robustness using San Francisco dictionary on Stanford Tower
  dataset.}
  \label{fig:robustness}
\end{figure}

Here we investigate, via numerical experiment, the robustness of the combination
of dictionary learning and SVD on different datasets from different
scenes. We first learn a
dictionary from the San Francisco data and apply SVD to it to
calculate a measurement matrix. This measurement matrix and dictionary is then used to compress and then reconstruct a different test data set, the Stanford
Tower dataset. We then perform the same test on the Stanford Tower
data, but this time using the dictionary and the corresponding
measurement matrix learned from the data
itself. Fig.~\ref{fig:robustness} shows relative error curves
of both approaches. As one can see, the relative error curve of
reconstruction using the
measurement matrix and the dictionary learned from the San Francisco data tested on the
Stanford Tower data (dashed red curve) is very close to the result
using the measurement matrix and the
dictionary learned from the Stanford Tower data applied to
itself (solid blue curve). In fact, the RMSE between these two error curves is
less than 0.11\%. This result indicates that our approach is
robust even when applied to different, but similar scenes.}

\section{Conclusion}\label{sec:conclusion}
In this paper, we proposed a new sampling strategy based on the existing
knowledge of the sparsifying dictionary of the spectral
signals we learned from the historical or training data to improve the
sampling efficiency and the
reconstruction performance for hyperspectral
imaging. Specifically,
a sparsifying dictionary was first learned from the training spectral data using
dictionary learning. We then used the first $m$ left singular vectors
as the fixed sampling pattern to obtain $m$ compressive
measurements. Then the spectral signals were reconstructed using the
SPGL1 solver based on the knowledge of the measurement matrix and the
sparsifying dictionary.
The reconstruction performance was further improved by reconditioning
the sensing matrix using matrix balancing, especially when the sensing
matrix became ill conditioned. \my{It was also shown that
our approach was robust to variations by applying the dictionary
learned from the training data and its associated singular vectors to
test data from different scenes.} 


%



\section*{Acknowledgment}

The authors would like to thank Dr. Stephen Gensemer and
Dr. Reza Arablouei for useful
discussions of the paper.

\ifCLASSOPTIONcaptionsoff
  \newpage
\fi



\bibliographystyle{IEEEtran}
\bibliography{ga}
\end{document}